# Rethinking the transfer learning for FCN based polyp segmentation in colonoscopy

Yan Wen[1], Lei Zhang[1*], Xiangli Meng[2] and Xujiong Ye[1]
[1]Laboratory of Vision Engineering (LoVE), School of computer science, University of Lincoln, LN6 7DQ, UK
[2]School of Electronic and Electrical engineering Lingnan Normal University, ZhanJiang, 524048, China

*Corresponding author: Lei Zhang (e-mail: lzhang@lincoln.ac.uk).

**ABSTRACT** Besides the complex nature of colonoscopy frames with intrinsic frame formation artefacts such as light reflections and the diversity of polyp types/shapes, the publicly available polyp segmentation training datasets are limited, small and imbalanced. In this case, the automated polyp segmentation using a deep neural network remains an open challenge due to the overfitting of training on small datasets. We proposed a simple yet effective polyp segmentation pipeline that couples the segmentation (FCN) and classification (CNN) tasks. We find the effectiveness of interactive weight transfer between dense and coarse vision tasks that mitigates the overfitting in learning. And It motivates us to design a new training scheme within our segmentation pipeline. Our method is evaluated on CVC-EndoSceneStill and Kvasir-SEG datasets. It achieves 4.34% and 5.70% Polyp-IoU improvements compared to the state-of-the-art methods on the EndoSceneStill and Kvasir-SEG datasets, respectively. The model and code are available at: https://github.com/MELSunny/Keras-FCN

## I. INTRODUCTION

Colorectal cancer is the fourth most common cancer in the UK, with over 42,000 new cases reported each year. Regular and accurate diagnosis is recommended to reduce mortality from colorectal cancer, especially for people in the age group of high risk. Screening is one of the effective methods for early detection and diagnosis of colorectal cancer. [1] During the screening operation, an optical colonoscopy is utilised to find colonic polyps, which are considered the early stage of colorectal cancer. Accurate detection of polyps and removal of them at the early stage can reduce the risk of colorectal cancer. [2] However, some studies have reported the miss rate of polyps during the screening is up to 28%. [3], [4]. In practice, the accuracy of detecting and monitoring polyps relies on the operator's experience in the colonoscopy operation.[5], [6] However, in some cases, it is challenging to distinguish polyps from the background since the visual characteristics of varied polyps are small [7], flat and sessile. [3] This, in turn, results in the missed detection of polyps during the screening. Moreover, human visual fatigue [8] is the additional factor leading to large intra-observer variabilities.

A series of methods have been proposed which have reached some success for automated polyp detection and segmentation, the success was mainly driven by the deep learning that has revolutionised computer vision over the last few years. At the core of these achievements are the significant advances of representation learning inherited in the CovNet with hierarchical network design.

The prior work leverages the CovNet to learn the complex appearances in colonoscopic frames, which includes intrinsic frame formation artefacts such as light reflections, bubbles and the diversity of polyp types and shapes. As the main limitation is due to the intrinsic data-hungry profile within deep learning methods, the size and scale of the dataset for training are crucial to model generalisation. However, the main challenge in the medical imaging field is the small dataset with limited samples and labels. More specifically, for the polyp segmentation and detection, to some extent, the accurate segmentation performance using deep networks relies on the large dataset and high-quality and consistent annotations/Ground truths. Nevertheless, on the one hand, the publicly available polyp segmentation datasets (~1 k images) are far smaller than the natural image datasets (1 million images) in the typical semantic segmentation vision task. On the other hand, it's very time-consuming to produce dense clinical annotations, and the consistency of the annotations is subject to the experts' experience. Therefore, beyond the diversity of the polyp appearance, the main challenge comes up with the small and imbalanced training samples, which is highly likely to lead to overfitting and training difficulty when training a deep network on a small dataset. In this case, we treat the polyp segmentation problem as developing a dense prediction method that works on the small dataset while mitigating the overfitting issue in the training process.

Transfer learning is one of the current well-known techniques to address the small dataset issue in training. It is crucial to boost the network performance in the prior segmentation methods, which could be categorised into two typical training pipelines regarding different initialisation strategies: 1) random initialisation with task-specific modules and 2) initialisation with pre-trained weights, e.g. ImageNet [9].

It is common to adopt random initialisation for those segmentation task-specific networks with customised layers or modules, where the main focus would be the design of new networks to lean more representative features. Whereas adopting well-established backbone networks with pre-training weights derived from a large-scale dataset is essential for the following finetuning on the target dataset, where the improvement is mainly gained from the simple weights transfer across the inter/intra domains.

In this study, given the small datasets in polyp segmentation, we followed the later strategy to investigate the efficiency of the new weights transfer scheme across domains that improves the segmentation while mitigating overfitting. We investigated the impact of transferring weights between two networks designed for different visual tasks (classification and segmentation). The two vision tasks were then integrated into a single segmentation pipeline sharing the finetuned weights.

More specifically, the Fully Convolution Network (FCN) with ResNet50 backbone and atrous convolution is used to generate the region proposals. Further, the classification convolution neural network with the same backbone is used for region proposal refinement. Specific transfer learning schemes are investigated, which are designed to mitigate overfitting by exploring the impact of inter-domain and intra-domain weights transfer in the learning stage.

Our main contributions can be summarised as follows:
1) We proposed a simple yet effective polyp segmentation pipeline that couples the segmentation (FCN) and classification (CNN) tasks. The atrous convolution was employed to enlarge the field of view for automatic polyp segmentation.
2) We proposed a new training scheme to train the network that mitigates the impact of overfitting in learning. We investigated the impact of transfer learning between two vision tasks. In our weight transfer scheme, we found the weights finetuned on the segmentation task (dense scale) can be used to accelerate the convergence in the classification task (coarse task) then the weights derived from the classification can be back-projected to the segmentation task to boost the segmentation performance continuously.
3) Our approach achieved state-of-the-art (SOTA) polyp segmentation performance on two publicly available datasets CVC-EndoSceneStill [10] and Kvasir-SEG dataset [11], with IoU of 76.68%, outperforming the previous state-of-the-art by 4.34% on the CVC-EndoSceneStill dataset. Notably, our method achieved the IoU of 80.22% on the Kvasir-SEG dataset, gaining a significant margin of 5.70% compared to the previous state of the art.

The rest of this paper is organised as follows. We summarise the related work in section II. Our method is described in section III, followed by the experimental results and analysis in section IV. The paper is concluded in section V.

**II. Related works**

A series of approaches for computer-aided polyp segmentation and detection have been proposed in colonoscopic videos over the last few years. The conventional methods proposed in earlier studies have commonly extracted intrinsic polyp features determined manually, such as texture information [12], colour [10], geometric features [13], [14] and contours [10].

In recent years, deep learning based methods have succussed and become the de-facto approach in multiple vision tasks and many deep learning based approaches have been proposed for polyp segmentation or detection in colonoscopy images. In comparison with traditional methods, deep learning based methods tend to have good segmentation performances. The fully convolutional network (FCN) [15] is the first end-to-end network for semantic segmentation using the convolutional network. David Vázquez et al [10] proposed a benchmark for endoluminal scene segmentation in Colonoscopy Frames and validated it on the FCN. They train and evaluate the fully convolutional network on the CVC-EndoSceneStill. By comparison, they found that FCN outperforms prior approaches in endoluminal scene segmentation. Another dataset Kvasir-SEG is proposed [11] and evaluated on the Deep Residual U-Net (ResUNet++) [16] as a baseline performance. Further, by applying Conditional Random Field and Test-Time Augmentation for ResUNet++ [17] Luisa F. Sánchez-Peralta et al [18] proposed a U-Net-based segmentation method and investigated the impacts of multiple image argumentation methods on two datasets, CVC-EndoSceneStill [10] and Kvasir-SEG. [11]They claimed that the most suitable transformation in data augmentation for each dataset is subject to the properties of the dataset, e.g., the polyp area, brightness and contrast. Further, they [19] present a new dataset for polyp detection, localisation and segmentation.

The combinations of two encoders (VGG16, Densenet121) and two decoders (LinkNet, VGG16) are evaluated on multiple datasets. Based on the best combination, the proposed method (LinkNet+ Densenet121) achieves the high accuracy on both CVC-EndoSceneStill [10] and Kvasir-SEG [11] test sets. The latest research [20] proposed a lightweight PolypSeg+. This model includes an adaptive scale context (ASC) module with a lightweight attention mechanism, and feature pyramid fusion (FPF).

In recent years, Vision Transformer [21] (ViT) has achieved highly competitive performance for image analysis tasks in multiple applications by splitting an image into patches and tokenizing the patches for feature extraction in Transformer. [22] proposed a novel ColonFormer, integrating Transformer and CNN as a unified architecture for polyp segmentation. A residual axial attention module is used for the refinement of segmentation. Another research [23] combines Res2Net [24] with ResNeXt [25] as a backbone, and uses Multi-Head self-attention and reverse attention mechanism for accurate polyp segmentation. In the field of transfer learning, research [26] on two large-scale datasets for medical imaging shows that transfer learning provides little performance improvement. However, we claim that ImageNet-pertained weights are critical to achieving competitive performance via weights transfer if the medical dataset is very small.

## III. Methods

The proposed automatic polyp segmentation framework consists of two convolutional neural networks. A fully convolutional network (FCN) [15] is utilised to segment polyps and a classification convolutional network (CNN) is employed to reduce the false positive rate. These two networks are designed by using the same backbone architecture ResNet50 [27] with atrous convolution [28]. A task-specific network training scheme via transfer learning is investigated to reduce the risk of overfitting in learning. Particularly, the effectiveness of sharing weights between the classification network and segmentation network (FCN) is investigated in this study.

### A. THE FRAMEWORK FOR POLYP SEGMENTATION

A general framework of the method is shown in Fig. 1. Our polyp segmentation framework is composed of two stages: Firstly, pixel-wise region proposals are generated by the fully convolutional network (FCN) with atrous convolution. Secondly, those region proposals are further extracted as the image patch candidates, which are classified into polyp/non-polyp patches by a binary CNN-based classifier.

### B. BACKBONE ARCHITECTURE

ResNet [27] is well known as a residual network. Compared to the previous convolution neural network (e.g. VGG19 [29]), it performs identity mapping and adds shortcut connections to simplify the optimisation and address the degradation problem without increasing the computational complexity. Thus, ResNet can benefit from gaining accuracy by increasing network depth. ResNet and its variant networks have shown remarkable performance in many downstream vision tasks, e.g., object classification/recognition and detection, semantic/instant segmentation, pose estimation and so on. In this experiment, we selected the plain 50 layers of the residual network (ResNet50) as the backbone structure for classification and segmentation considering the size of the training samples.

### C. ATROUS CONVOLUTION

Atrous convolution is a kind of special convolution with a sparse convolution kernel. The input signal is sampled with interspace. The dilate rate in atrous convolution defines the size of the interspace. It allows the network to enlarge the field of view of filters to incorporate a larger context with reduced computation. It controls the field-of-view and balances the trade-off between accurate localisation (small field-of-view) and context assimilation (large field-of-view).

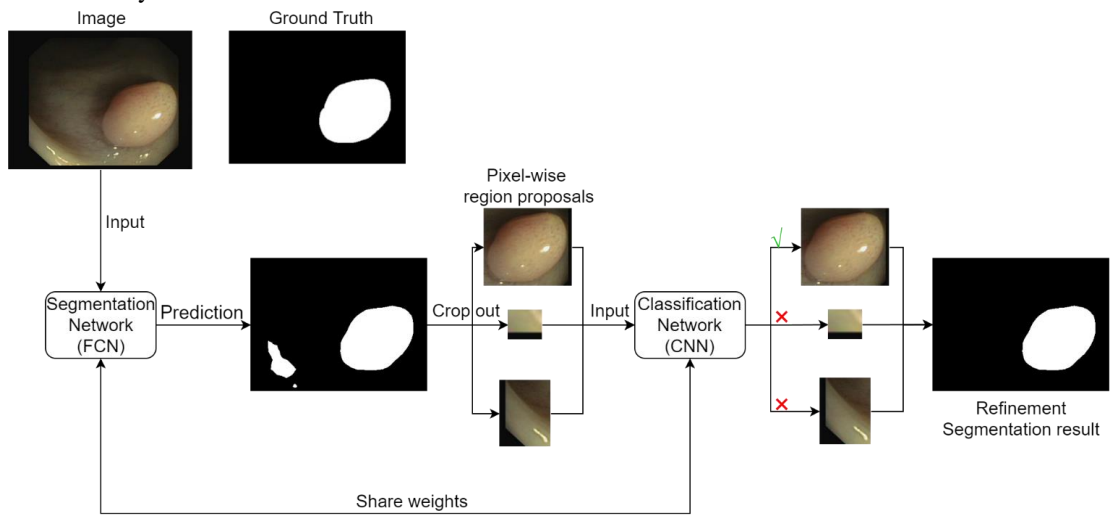

**FIGURE 1.** The framework of the method. The input image is initially segmented by Fully Convolution Network. The candidate patches are further classified by the Classification network using CNN. The result's false-positive objects are removed finally.

In our network, instead of using pooling layers that could potentially reduce the resolution of the feature map, the atrous convolution is employed to enlarge the field of view (FOV) and keep the resolution of the feature map with even less computation.

The overview backbone structure of our FCN is presented in Fig. 2. Each identity block contains 3 stacked convolution layers whose kernel sizes are 1×1, 3×3 and 1×1, respectively as well as the identity shortcuts connections. A slightly different design compared with the original ResNet50, the last three block uses a 2×2 atrous convolution with a 1×1 stride to avoid downsampling of feature size.

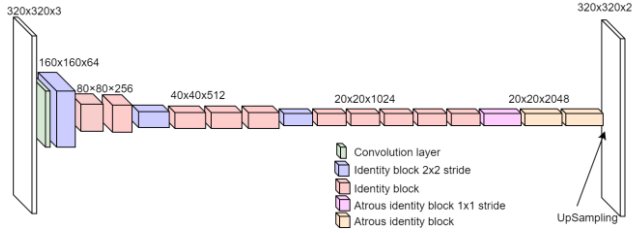

**FIGURE 2.** The overview structure of our segmentation FCN network with atrous convolution.

### D. CNN-BASED REGION REFINEMENT

The region refinement is the postprocessing based on the initial prediction from the FCN. More specifically, the refinement firstly utilises the find-contours function [30] to distinguish the multiple candidates of objects from the label maps of the segmentation. For each candidate of objects, the image patches are cropped out by the bounding box of the contours. Each cropped image patch is then fed to the patch classification network (Fig. 3) for refinement. This design was based on the observation that the foreground objects (polyps) are rather smaller than the background but similar to other artefacts, which could lead to some false positives. In our method, the region refinement is conducted by patch-based classification, considering the features at different scales (pixel-wise against patch-wise). As we use the same resnet50 as the backbone in classification and segmentation, the weights trained from different vision tasks can mutually be transferred to each other. For example, the pre-trained weights can transfer to the classification network so that the final segmentation can be enhanced by recognising and removing false positive objects from the initial segmentation.

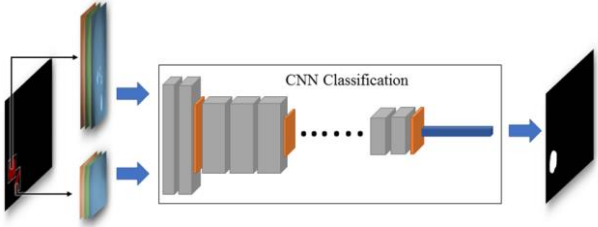

**FIGURE 3.** The CNN-based region refinement. The inference from FCN has 2 objects and these objects are cropped into image patches. Each image patch is classified by CNN. Based on the result of CNN, the false positive objects are removed.

### E. TRAINING SCHEME

The proposed segmentation method was trained using the following protocol (Fig. 4 training scheme 1). Firstly, the FCN is initialised with the ImageNet pre-trained weights and finetuned on the colonoscopy training dataset. Secondly, the finetuned weights of the FCN are transferred to the classification CNN network with the same resnet50 backbone to refine the polyp patches. Finally, the tuned weights of reset50 on the CNN classification are circulated to FCN again for the final segmentation tuning. We also conducted an alternative training loop, which starts from CNN classification followed by FCN tunning and ends with classification tunning again (Fig. 4 training scheme 2). We empirically found that training scheme1 can mitigate overfitting in training and continually boost the FCN performance. More specifically, weight transfer from a dense task (segmentation) to a coarse task (classification) can accelerate the CNN training convergence. Meanwhile, it is worth noting that the FCN segmentation performance can be continually improved by circulating the finetuned weights of CNN back to the FCN. More experimental results and comparisons are presented in section IV.C *ablation study*.

### F. LOSS FUNCTIONS

For training the classification networks, Binary Cross Entropy loss is used for the optimisation:

$$Loss_{BE}(y, \hat{y}) = -\frac{\sum_{n=1}^{N} \hat{y}_n \times \log y_n + (1-\hat{y}_n) \log(1-y_n)}{N} \quad (1)$$

Where, $y_n$ is the $n$th probability in the model output, $\hat{y}_n$ is the $n$th value in the corresponding ground truth. $N$ is the minibatch size.

For evaluation of the segmentation models, the similar Softmax Sparse Cross Entropy loss is used:

$$Loss_{CE}(y, \hat{y}) = -\frac{\sum_{n=1}^{N} \hat{y}_n \times \log \sigma(y_n)}{N} \quad (2)$$

$$\sigma(x)_i = \frac{e^{x_i}}{\sum_{j=1}^{C} e^{x_j}} \quad (3)$$

Where $\sigma$ is the softmax function, making the label map $\hat{y}$ normalised to 0~1, meanwhile, for each class on a specific voxel, the total of all class values is 1. $x_i$ is the feature prediction for class $i$, $C$ is the total amount of classes.

### G. EVALUATION METRIC

A set of standard classification, segmentation and detection evaluation metrics are employed to evaluate the method. The classification performance is evaluated by the accuracy metric, the segmentation is evaluated by the IoU also known as the Jaccard index.

**Classification**

$$Accuracy = \frac{TP}{TP+FP} \quad (4)$$

## Segmentation

$$IoU(X,Y) = \frac{\sum_{c=1}^{N} \frac{|X_c \cap Y_c|}{|X_c \cup Y_c|}}{N} = \frac{\sum_{c=1}^{N} \frac{TP_c}{TP_c + FP_c + FN_c}}{N} \quad (5)$$

$$Dice(X,Y) = \frac{\sum_{c=1}^{N} \frac{2|X_c \cap Y_c|}{|X_c| + |Y_c|}}{N} = \frac{\sum_{c=1}^{N} \frac{2TP_c}{2TP_c + FP_c + FN_c}}{N} \quad (6)$$

$$Mean\ IoU = \frac{IoU_{Polyp} + IoU_{Background}}{2} \quad (7)$$

Where $N$ is the number of classes. $|X_c|$ and $|Y_c|$ are the number of voxels for class $c$, $TP_c$ is the true positive for class $c$, $FP_c$ is false positive for class $c$, $FN_c$ is the false negative for class $c$

## IV. EXPERIMENTS AND RESULTS

### A. EXPERIMENT MATERIALS

To validate our method and training scheme, two public datasets (the CVC-EndoSceneStill [10] and the Kvasir-SEG datasets [11]) are used for training and evaluation in our experiments.

The CVC-EndoSceneStill is a combination of CVC-ColonDB and CVC-ClinicDB, which includes 912 images obtained from 44 video sequences acquired from 36 patients. In the CVC-ColonDB, there are 300 colonoscopy images with a resolution of 500 × 574 pixels. And the CVC-ClinicDB has 612 colonoscopy images with a lower resolution of 384 × 288 pixels. According to the metadata, the dataset is split into 547 training frames, 183 validation frames and 182 testing frames.

The Kvasir-SEG dataset [11] has relatively more images which contain 1000 colonoscopy frames of various sizes. The image resolution ranges from 332 × 487 to 1920 × 1072 pixels. We followed the same strategy in [20] to split the dataset into training, validation and tests for comparison. More specifically, the total of 1000 images is split into 800 training images, 100 validation images and 100 testing images.

In the experiments, the binary label is used for training and evaluation of polyp patch classification, while for the segmentation, two classes are used, where pixels corresponding to the polyp are labelled with 1 and 0 otherwise.

### B. TRAINING PROTOCOL AND SETTINGS

Two sets of training hyperparameters are applied to the classification and segmentation network, respectively. The details of the settings are summarized as follows.

Classification network:
- SGD optimiser with 0.9 momentum
- 1e-3 initial learning rate with cosine decay
- 250 training epochs
- 24 batch size

Segmentation network:
- Adam optimiser [31] with amsgrad [32]
- 1e-4 learning rate
- 500 training epochs
- 24 batch size

We developed an image patch generator to generate image patches (polyp or non-polyp patch) for the classification network training. Specifically, image patches of polyp are generated by a bounding box on the polyp of the segmentation mask. And then, background image patches are generated by random cropping on the image but excluding the polyp region.

Since the dataset is small, with only 547 training images, data augmentation was employed in our experimentations that could potentially reduce the overfitting, and the following data augmentation factors were applied to increase the training simples:
- Keep aspect ratio resize from range 0.8x to 1.25x with padding zero value
- 0° to 180° rotation range
- Random horizontal and vertical flip
- 0.1 Width and height Shift range

All experiments were conducted on the machine with six core 12 threading CPUs running on 3.8Ghz (Intel Core i7 6850K) with 32GB DDR4 RAM. The NVIDIA RTX 3090 GPU with 24GB VRAM is used to accelerate the training. The implementation is based on TensorFlow [33] and Keras. The software development environment is Ubuntu 20.04 with NVIDIA CUDA 11.

### C. ABLATION STUDY
**Input size**

We argue that the input size affects the overall performance of the polyp segmentation. We validated various input sizes of 192×192, 224×224, 256×256, 320×320 and 384×384 in training and validation to spot an optimised input size for the segmentation. The commonly used metric of polyp Intersection over Union (IoU) is accessed for these segmentation networks with different input sizes. The result in Fig. 5 shows that the input size of 320×320 has reached the highest performance for the segmentation in the validation set. Compared to the size of 384 × 384, we can observe that the size of 320 × 320 input has very competitive performance in the early training epochs but has better IoU since 160 epochs. This counterintuitively shows that the segmentation performance won't keep improving when increasing the input size in the FCN network. In our experiments, we adopted the input size of 320 ×320 for segmentation, considering its superior performance as well as the slightly less computational cost.

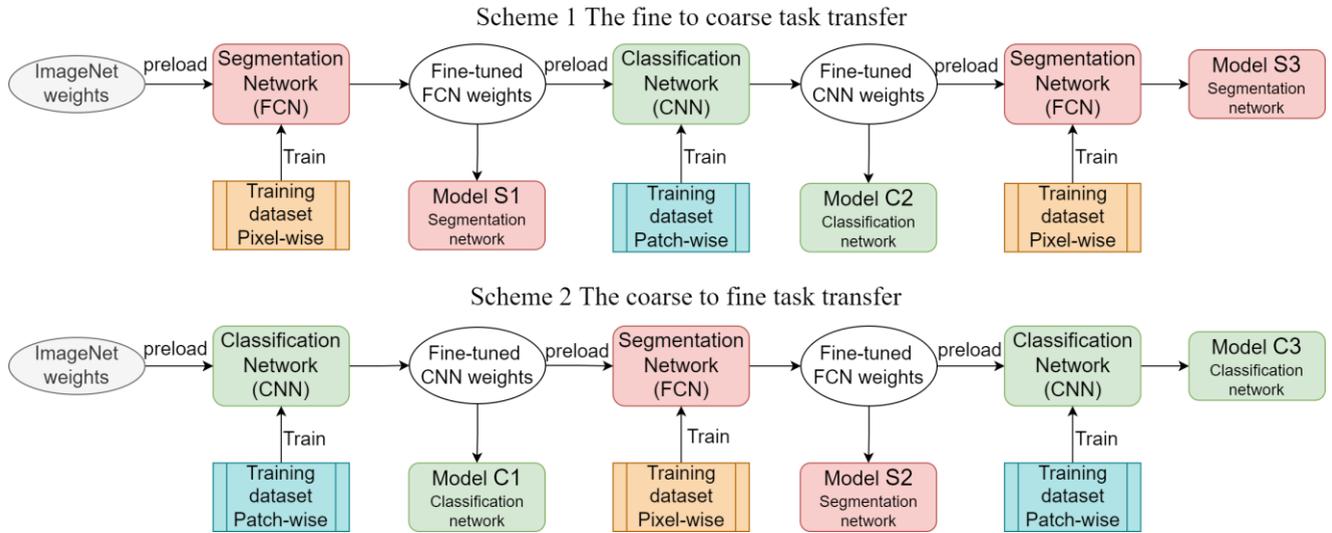

**FIGURE 4.** Two training schemes, The ImageNet weights in scheme 1 on the top, start with finetuning on FCN, then are trained on CNN and finally return to FCN for final finetuning. In scheme 2, the ImageNet weights start with training on CNN, and so on.

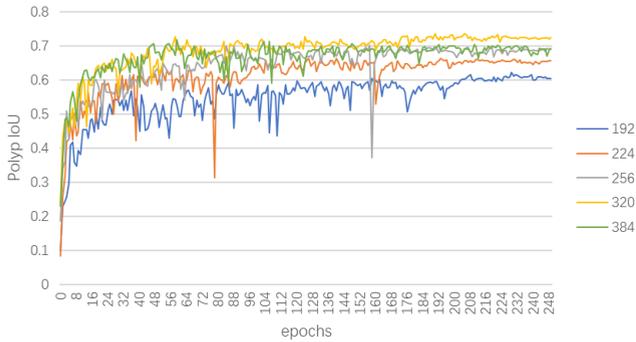

**FIGURE 5.** Evaluation for segmentation networks with various input sizes. The numbers in the legend indicate the input size.

**Training Scheme and weights transfer**

We investigated the performance and impact of sharing weights between segmentation and classification networks, which are equipped with the same backbone architecture (ResNet 50). Two training schemes shown in Fig. 4 for the proposed method are conducted to investigate the impact of the weights transferring between Patch-wise coarse prediction (classification) and pixel-wise dense prediction (segmentation). In the first scheme (i.e., top flow in Fig. 4 FCN->CNN->FCN), the weights are transferred in the path of learning from dense prediction task (with FCN) to patch-wise classification task (CNN), and then we circulated the trained CNN weights back to FCN again. More specifically, we finetuned the FCN Segmentation (noted as Model S1) on the polyp dataset, of which the backbone was initialized by ImageNet pre-trained weights. And then, we finetuned the CNN classification network (Model C2) by preloading weights of FCN (Model S1). Finally, the finetuned CNN weights were transferred back to the FCN again for the final segmentation training, denoted as FCN (Model S3).

Alternatively, in the second scheme (i.e., bottom flow in Fig. 4 corresponding to the path CCN->FCN->CCN as coarse to fine task transfer), We initialised CNN (Model C1) with ImageNet pre-trained weights and trained it on the colon patches, then we finetuned FCN (Model S2) by preloading CNN (Model C1) weights. Finally, we finetuned CNN (Model C3), with preloading FCN (Model S2) weights.

Additionally, to validate the weights transfer with or without pre-trained weights, the random weights initialisation for the segmentation FCN (denoted as Model S0) is used in training for the comparison. This ablation is also designed to assess the effectiveness of weight transfers from the natural to the medical domain.

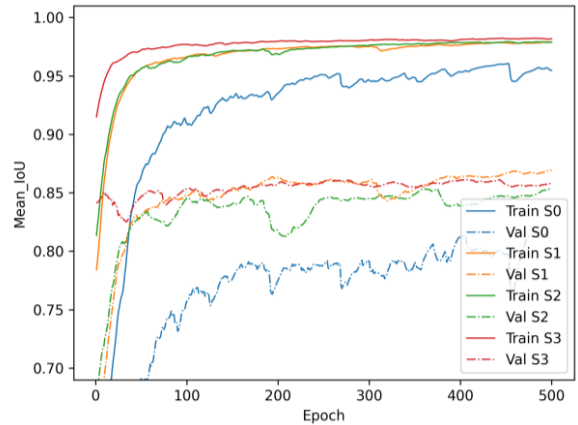

**FIGURE 6.** Evaluations of the segmentation networks, Dashed lines denote validation IoU, solid lines denote training IoU

We evaluated two training schemes on the polyp segmentation dataset. Fig. 6 shows the training and validation curves of all FCN networks (S0-S3) with different initialisation strategies with or without transfer learning. We can observe that model (S0) training with the randomly initialised weights has slower convergence compared to the other models (S1-S3), which are initialised or finetuned based on the ImageNet pre-training weight. The S0 also tends to be overfitting more easily. We conjecture that on the small dataset, the inter-domain or cross-domain weight transfer from the natural images ImageNet to the medical image domain (polyp images) is non-trivial for the downstream polyp segmentation task.

Moreover, Fig. 6 shows that the S1 of training scheme 1 has a similar convergence in training compared to S2 of the training scheme2. However, the fluctuance of the S2 validation (Val 2) curve demonstrates unstable performance and less generalisation. This indicates that fine-tuning with coarse classification across domains before dense segmentation does not help the segmentation. We speculate the weights directly transferred from the ImageNet to the polyp domain are corrupted during the patch-wise classification finetuning due to the limited training samples. However, the classification finetuning (C2) initialised by the intra-domain tuned weights from S1 has a stepstone effect that continuously boosts the segmentation performance, as shown in Fig. 6. model S3. The training and validation curves of S3 demonstrate faster convergence in the early epochs and training stability in the ongoing epochs.

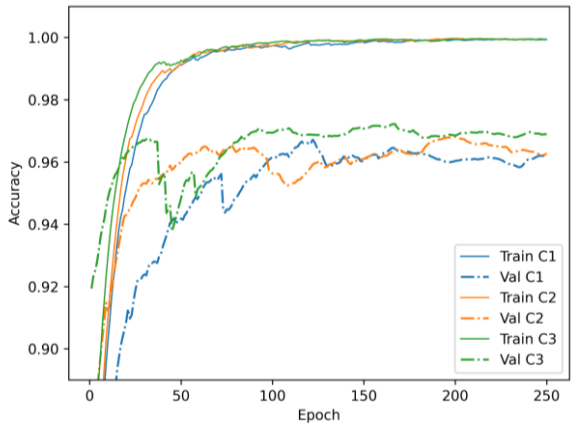

**FIGURE 7.** Evaluations of the classification networks,
Dashed lines denote validation accuracy, solid lines denote training accuracy

We further investigate the effectiveness of the weight transfer in classification networks within two training schemes. Fig. 7 presents the evaluation results in terms of accuracy for the classification of the validation dataset. We can observe that model C1 using ImageNet weights has a slower convergence speed compared to model C3 with the weights of model S2. Unsurprisingly, this indicates the positive impact on the transfer of intra-domain weights from the dense (segmentation) to the coarse prediction task (classification). Although model C3 in scheme 2 has a slightly improved performance and convergence speed than model C2 in scheme 1 in both training and validation phases, the C3 tends to be unstable with large fluctuation in validation around 50 epochs in Fig. 7. The competitive performance between the C2 and C3 (both are initialised with FCN-trained weights) also verifies the benefits of transferring the weight of the segmentation task to classification.

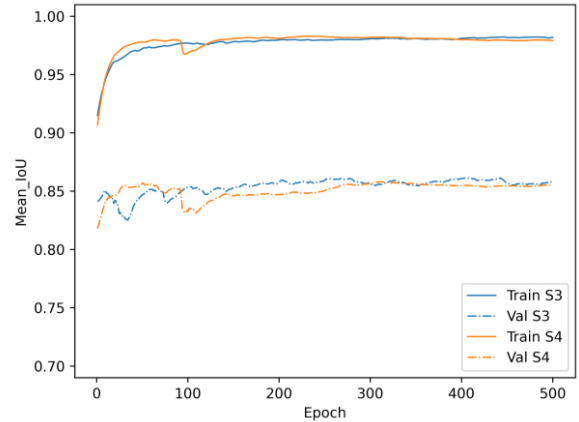

**FIGURE 8.** Evaluations of the classification networks,
Dashed lines denote validation accuracy, solid lines denote training accuracy

In addition, to verify whether FCN can be further improved by the aforementioned iterative training scheme 1 (FCN-CNN-FCN), We stacked one more iteration, namely (FCN-CNN-FCN-CNN-FCN), sharing weights between FCN and CNN, denoted the model S4. Fig. 8 shows that model S4 does not have significant improvements. Instead, the S3 demonstrates slightly faster convergence. This indicates that the performance of FCN cannot be continuously improved by sharing weights between segmentation and classification networks.

Based on our ablations and discovered components leading to the segmentation improvement, we choose training scheme 1 with models S3 and C2 for polyp segmentation.

### D. SEGMENTATION EVALUATION RESULT

TABLE I
COMPARISON OF THE SEGMENTATION METHODS ON THE CVC-ENDOSCENESTILL TEST DATASET

| Methods | Polyp IoU | Polyp Dice | mIoU | Pixel Accuracy |
|---|---|---|---|---|
| U-Net | 65.33% | 75.28% | | 93.24% |
| FCN-8s [10] | 50.85% | | 72.74% | 94.91% |
| FCN-8s [34] | 66.99% | 75.53% | | 94.03% |
| U-Net + Dilated Conv [35] | 58.60% | 67.32% | | 94.12% |
| PolypSeg+ [20] | 63.50% | 71.70% | | 94.93% |
| LinkNet + Densenet121 [19, p.] | 72.16% | | | |
| U-Net + Transformations [18] | 72.34% | | | |
| EFCN-8 [36] | 58.7% | | 76.7% | 94.9% |
| PraNet [37] | | 79.7% | | |
| ColonFormer-L [22] | | | 84.2% | |
| Our-S1 (FCN16s-Atrous) | 75.84% | 86.20% | 86.2% | 96.82% |
| Our-S3 | 76.58% | 86.74% | 86.60% | 96.87% |
| **Our-S3+C2 Refinement** | **76.68%** | **86.80%** | **86.66%** | **96.88%** |

Our segmentation method is evaluated on two public datasets (CVC-EndoSceneStill and Kvasir-SEG). The evaluation results, alongside the previous methods on the CVC-EndoSceneStill test set, are summarized in Table I. The S1 model achieves 75.84% of polyp IoU, which verifies the improvement of polyp features by introducing the Atrous convolution in the FCN-16s backbone. The segmentation performances are further improved by adopting our training scheme that achieves 76.58% of Polyp IoU and 86.74% of dice scores with the same or fewer computations. With the refinement (S3+C2), our method achieves the IoU of 76.68%, outperforming the previous state-of-the-art by 4.34% on the CVC-EndoSceneStill dataset. It's worth noting that our pure convolution-based method outperforms the latest transformer-based method and achieves 86.66% of mIoU on the EndoSceneStill test dataset.

TABLE II
REFINEMENT ON THE CVC-ENDOSCENESTILL TEST DATASET

| Methods | TP | FP | FN |
|---|---|---|---|
| S3 | 159 | 88 | 23 |
| S3+C2 | 158 | 64 | 24 |

As the combination of the S3+C2 has further improvement compared with S3, we assess the effectiveness of the refinement and report the results in table II. TABLE II shows that some false positive objects are removed which improves the polyp segmentations e.g., IoU, and accuracy. Fig. 9 illustrates examples that show how refinement improved the segmentation inference. The initial segmentations derived from S0 (Fig. 9 Column c) contain a few segmented objects, some of which are false positives. These false positives are removed after the refinement by polyp patch classification, whilst the true positives in examples remain.

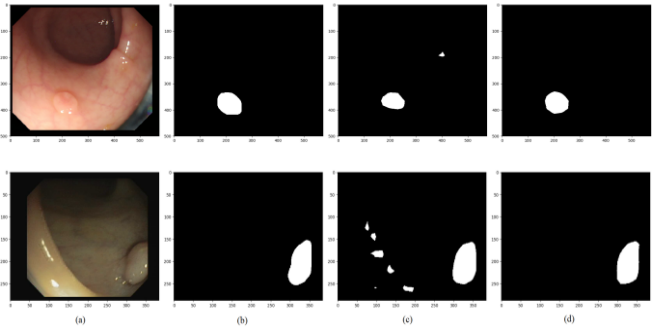

**FIGURE 9.** Comparison of Segmentation results. (a) Input image (b) Ground Truth (c) FCN with ImageNet weights initialization (d) the proposed method

We find consistent improvements in our method on the Kvasir-SEG dataset that shows the effectiveness of the proposed method and training scheme. On average, mode S3+C2 achieves significant 2.48% polyp IoU improvements compared to model S1. Our method achieves state-of-the-art performance on this dataset in terms of metrics Polyp IOU, Polyp Dice, mean IOU and Accuracy with 80.22%, 89.02%, 87.97% and 96.38%, respectively. Notably, the polyp IoU of 80.22% on the Kvasir-SEG dataset, gaining a significant margin of 5.70% compared to the previous state of the art.

TABLE III
COMPARISON OF THE SEGMENTATION METHODS ON THE KVASIR-SEG TEST DATASET

| Methods | Polyp IoU | Polyp Dice | mIoU | Pixel Accuracy |
|---|---|---|---|---|
| U-Net | 65.33% | 75.28% | | 93.24% |
| FCN-8s [34] | 66.05% | 76.02% | | 93.35% |
| U-Net + Dilated Conv [35] | 66.54% | 76.06% | | 93.45% |
| PolypSeg+ [20] | 70.68% | 78.81% | | 94.08% |
| LinkNet + Densenet121 [19, p.] | 74.52% | | | |
| U-Net + Transformations [18] | 68.17% | | | |
| Trans-PraNeXt [23] | | | 84.6% | |
| ColonFormer-L [22] | | | 87.6% | |
| Our-S1 (FCN16s-Atrous) | 77.74% | 87.47% | 86.31% | 96.82% |
| Our-S3 | 80.15% | 88.98% | 87.92% | 96.37% |
| **Our-S3+C2 Refinement** | **80.22%** | **89.02%** | **87.97%** | **96.38%** |

Fig. 10 shows the segmentation results of our method (S3 + C2 refinement) on some representative cases, including polyps in various sizes and shape, as well as, various image attributes, including resolution, clarity, and brightness. Thanks to the atrous convolution, the receptive field of the model is widened, as a result, the polyps with large scale are segmented effectively in the result of (a) and (d). More challenging cases of some imperceptible polyps, caused by the confusion with folds (g), low clarity image (e) and edge location, are successfully segmented. In the result of (g), the proposed method gives a larger segmented region than the ground truth. But the uncertainty remains in deciding the precise polyp region in this case. The large area from the model inference can be preserved for further analysis by the radiologist. The result of (c) shows that the method has the ability of multi-object segmentation. Case (i) contains an irregularly shaped polyp, and the proposed method provides a consistent contour to the ground truth.

Regarding efficiency, TABLE IV summarises the inference time of the segmentation approaches on colonoscopy. Our model S3 approach is 4.35 milliseconds, achieving an average of 23 FPS on the single GPU card, which is sufficient for real-time application. The postprocessing costs extra 2.76 milliseconds, resulting average of 14 FPS. The speed of our approach is close to the fastest [20], while S3 cost additional 1.13 milliseconds achieving 13% more on polyp IoU. Moreover, its inference is based on batch size 1 with single CPU threading for pre-processing and post-processing. Lower latency and higher fps could be achieved after proper optimisation.

TABLE IV
PROCESSING TIME FOR SEGMENTATION

| Methods | Processing time | Polyp IoU |
|---|---|---|
| **S3+C2** | **7.11ms** | **76.68%** |
| S3 | 4.35ms | 76.58% |
| FCN-8s [10] | 88ms | 50.85% |
| U-Net + Dilated Conv [35] | 50ms | 58.60% |
| PolypSeg+ [20] | 3.22ms | 63.50% |

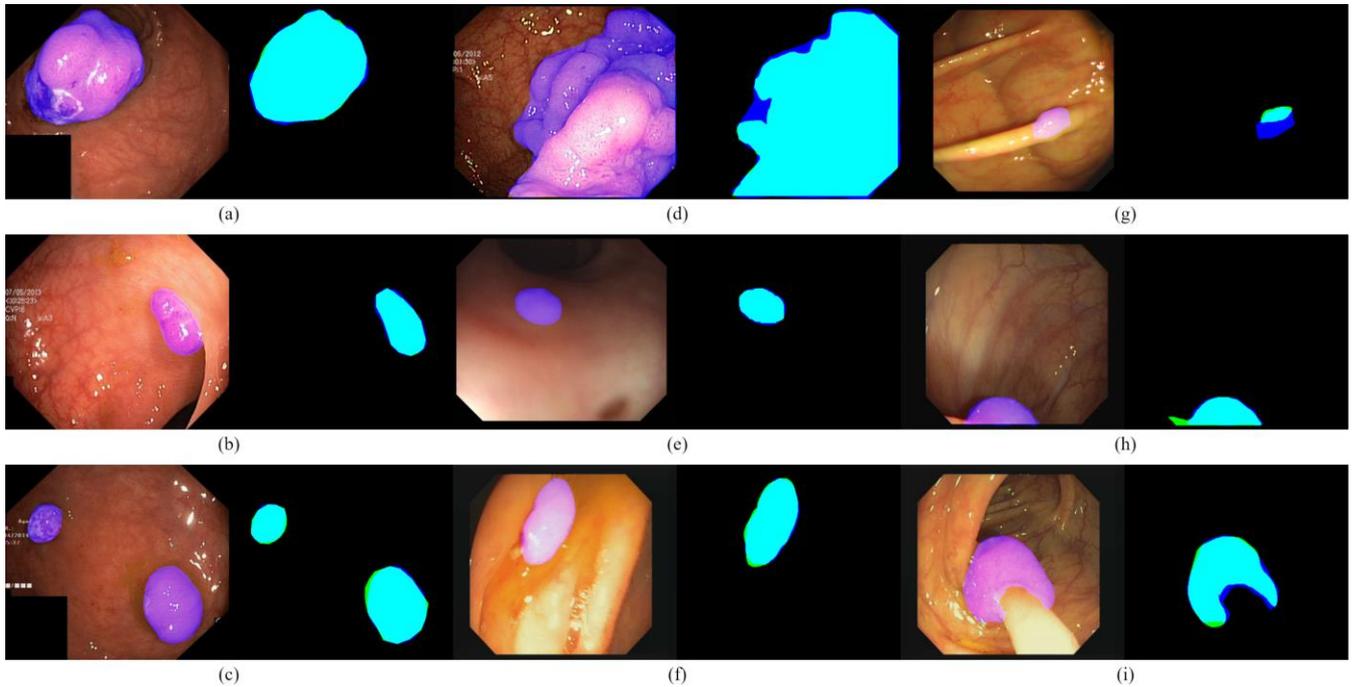

**FIGURE 10.** Qualitative results. (a)~(d) are cases from the Kvasir-SEG test set, (e)~(i) are cases from CVC-EndoSceneStill. In each case, the left image is the input image with highlighted (purple) prediction generated from our method. The right image is the overlap (aqua) of ground truth(green) and prediction (blue).

## V. CONCLUSION

In this paper, we propose an automatic polyp segmentation method in colonoscopy frames which integrates atrous FCN with ResNet50 backbone for region proposal and classification CNN for region refinement. A new training scheme is designed to mitigate overfitting due to the small dataset, which is a common issue in medical imaging. We discover several key components regarding the weight transfer leading to the improvement of polyp segmentation. We find that circulating the weight transfer between dense and coarse prediction tasks (segmentation and classification) can mitigate overfitting even using the small dataset and continue to boost the final segmentations. This motivates us to design a segmentation framework with a new training scheme. Our method is evaluated on two pubic datasets (CVC-EndoSceneStill and Kvasir -SEG). It achieves the segmentation performance with 86.80% of Polyp Dice, 86.66% of mean IOU and 96.88%of Accuracy on the EndoSceneStill dataset, and 89.02% of Polyp Dice, 87.97% of mean IOU and 96.38%of Accuracy on the Kvasir -SEG. Our method achieves state-of-the-art and outperforms the existing methods on polyp segmentation with the same or fewer computations by 4.34% and 5.70% of Polyp IoU on EndoSceneStill and Kvasir-SEG datasets, respectively.

## APPENDIX

The implantation of our experiment has been released on GitHub: https://github.com/MELSunny/Keras-FCN

## ACKNOWLEDGMENT

This work was partially supported by British Council UK-ASEAN Institutional Links ECR Scheme (Real-Time Data Quality Analysis and Control for Aquaculture Prawn Farming Management) project (913100317).

[8] P. Liu *et al.*, 'The single-monitor trial: an embedded CADe system increased adenoma detection during colonoscopy: a prospective randomized study', *Therap Adv Gastroenterol*, vol. 13, p. 1756284820979165, Dec. 2020, doi: 10.1177/1756284820979165.

[9] J. Deng, W. Dong, R. Socher, L.-J. Li, K. Li, and L. Fei-Fei, 'ImageNet: A large-scale hierarchical image database', in *2009 IEEE Conference on Computer Vision and Pattern Recognition*, Jun. 2009, pp. 248–255. doi: 10.1109/CVPR.2009.5206848.

[10] D. Vázquez *et al.*, 'A Benchmark for Endoluminal Scene Segmentation of Colonoscopy Images', *J Healthc Eng*, vol. 2017, p. 4037190, 2017, doi: 10.1155/2017/4037190.

[11] D. Jha *et al.*, 'Kvasir-SEG: A Segmented Polyp Dataset', in *MultiMedia Modeling*, vol. 11962, Y. M. Ro, W.-H. Cheng, J. Kim, W.-T. Chu, P. Cui, J.-W. Choi, M.-C. Hu, and W. De Neve, Eds. Cham: Springer International Publishing, 2020, pp. 451–462. doi: 10.1007/978-3-030-37734-2_37.

[12] A. V. Mamonov, I. N. Figueiredo, P. N. Figueiredo, and Y.-H. Richard Tsai, 'Automated Polyp Detection in Colon Capsule Endoscopy', *IEEE Transactions on Medical Imaging*, vol. 33, no. 7, pp. 1488–1502, Jul. 2014, doi: 10.1109/TMI.2014.2314959.

[13] S. Hwang, J. Oh, W. Tavanapong, J. Wong, and P. C. de Groen, 'Polyp Detection in Colonoscopy Video using Elliptical Shape Feature', in *2007 IEEE International Conference on Image Processing*, Sep. 2007, vol. 2, p. II-465-II-468. doi: 10.1109/ICIP.2007.4379193.

[14] J.-G. Lee, J. Hyo Kim, S. Hyung Kim, H. Sun Park, and B. Ihn Choi, 'A straightforward approach to computer-aided polyp detection using a polyp-specific volumetric feature in CT colonography', *Comput Biol Med*, vol. 41, no. 9, pp. 790–801, Sep. 2011, doi: 10.1016/j.compbiomed.2011.06.015.

[15] J. Long, E. Shelhamer, and T. Darrell, 'Fully convolutional networks for semantic segmentation', in *2015 IEEE Conference on Computer Vision and Pattern Recognition (CVPR)*, Jun. 2015, pp. 3431–3440. doi: 10.1109/CVPR.2015.7298965.

[16] Z. Zhang, Q. Liu, and Y. Wang, 'Road Extraction by Deep Residual U-Net', *IEEE Geoscience and Remote Sensing Letters*, vol. 15, no. 5, pp. 749–753, May 2018, doi: 10.1109/LGRS.2018.2802944.

[17] D. Jha *et al.*, 'A Comprehensive Study on Colorectal Polyp Segmentation With ResUNet++, Conditional Random Field and Test-Time Augmentation', *IEEE Journal of Biomedical and Health Informatics*, vol. 25, no. 6, pp. 2029–2040, Jun. 2021, doi: 10.1109/JBHI.2021.3049304.

[18] L. F. Sánchez-Peralta, A. Picón, F. M. Sánchez-Margallo, and J. B. Pagador, 'Unravelling the effect of data augmentation transformations in polyp segmentation', *Int J CARS*, vol. 15, no. 12, pp. 1975–1988, Dec. 2020, doi: 10.1007/s11548-020-02262-4.

[19] L. F. Sánchez-Peralta *et al.*, 'PICCOLO White-Light and Narrow-Band Imaging Colonoscopic Dataset: A Performance Comparative of Models and Datasets', *Applied Sciences*, vol. 10, no. 23, Art. no. 23, Jan. 2020, doi: 10.3390/app10238501.

[20] H. Wu, Z. Zhao, J. Zhong, W. Wang, Z. Wen, and J. Qin, 'PolypSeg+: A Lightweight Context-Aware Network for Real-Time Polyp Segmentation', *IEEE Transactions on Cybernetics*, pp. 1–12, 2022, doi: 10.1109/TCYB.2022.3162873.

[21] A. Dosovitskiy *et al.*, 'An Image is Worth 16x16 Words: Transformers for Image Recognition at Scale'. arXiv, Jun. 03, 2021. Accessed: Aug. 14, 2022. [Online]. Available: http://arxiv.org/abs/2010.11929

[22] N. T. Duc, N. T. Oanh, N. T. Thuy, T. M. Triet, and V. S. Dinh, 'ColonFormer: An Efficient Transformer Based Method for Colon Polyp Segmentation', *IEEE Access*, vol. 10, pp. 80575–80586, 2022, doi: 10.1109/ACCESS.2022.3195241.

[23] X. Zhou, Z. Bai, Q. Lu, A. Huang, and S. Fan, 'Colorectal Polyp Segmentation Based on Group Convolution and Transformer', in *2021 China Automation Congress (CAC)*, Oct. 2021, pp. 6387–6392. doi: 10.1109/CAC53003.2021.9728295.

[24] S.-H. Gao, M.-M. Cheng, K. Zhao, X.-Y. Zhang, M.-H. Yang, and P. Torr, 'Res2Net: A New Multi-Scale Backbone Architecture', *IEEE Transactions on Pattern Analysis and Machine Intelligence*, vol. 43, no. 2, pp. 652–662, Feb. 2021, doi: 10.1109/TPAMI.2019.2938758.

[25] S. Xie, R. Girshick, P. Dollar, Z. Tu, and K. He, 'Aggregated Residual Transformations for Deep Neural Networks', in *2017 IEEE Conference on Computer Vision and Pattern Recognition (CVPR)*, Honolulu, HI, Jul. 2017, pp. 5987–5995. doi: 10.1109/CVPR.2017.634.

[26] M. Raghu, C. Zhang, J. Kleinberg, and S. Bengio, 'Transfusion: Understanding Transfer Learning for Medical Imaging', in *Advances in Neural Information Processing Systems*, 2019, vol. 32. Accessed: Sep. 25, 2022. [Online]. Available: https://proceedings.neurips.cc/paper/2019/hash/eb1e78328c46506b46a4ac4a1e378b91-Abstract.html

[27] K. He, X. Zhang, S. Ren, and J. Sun, 'Deep Residual Learning for Image Recognition', in *2016 IEEE Conference on Computer Vision and Pattern Recognition (CVPR)*, Las Vegas, NV, USA, Jun. 2016, pp. 770–778. doi: 10.1109/CVPR.2016.90.

[28] L.-C. Chen, G. Papandreou, I. Kokkinos, K. Murphy, and A. L. Yuille, 'DeepLab: Semantic Image Segmentation with Deep Convolutional Nets, Atrous Convolution, and Fully Connected CRFs'. arXiv, May 11, 2017. Accessed: Aug. 22, 2022. [Online]. Available: http://arxiv.org/abs/1606.00915

[29] K. Simonyan and A. Zisserman, 'Very Deep Convolutional Networks for Large-Scale Image Recognition'. arXiv, Apr. 10, 2015. doi: 10.48550/arXiv.1409.1556.

[30] 'scikit-image: Image processing in Python'. Image Processing Toolbox for SciPy, Aug. 31, 2022. Accessed: Aug. 31, 2022. [Online]. Available: https://github.com/scikit-image/scikit-image/blob/00177e14097237ef20ed3141ed454bc81b308f82/skimage/measure/_find_contours.py

[31] D. P. Kingma and J. Ba, 'Adam: A Method for Stochastic Optimization'. arXiv, Jan. 29, 2017. doi: 10.48550/arXiv.1412.6980.

[32] S. J. Reddi, S. Kale, and S. Kumar, 'On the Convergence of Adam and Beyond'. arXiv, Apr. 19, 2019. Accessed: Aug. 30, 2022. [Online]. Available: http://arxiv.org/abs/1904.09237

[33] Paszke A. *et al.*, 'Automatic differentiation in PyTorch', Oct. 2017, Accessed: Aug. 26, 2022. [Online]. Available: https://openreview.net/forum?id=BJJsrmfCZ

[34] M. Akbari *et al.*, 'Polyp Segmentation in Colonoscopy Images Using Fully Convolutional Network', in *2018 40th Annual International Conference of the IEEE Engineering in Medicine and Biology Society (EMBC)*, Honolulu, HI, Jul. 2018, pp. 69–72. doi: 10.1109/EMBC.2018.8512197.

[35] X. Sun, P. Zhang, D. Wang, Y. Cao, and B. Liu, 'Colorectal Polyp Segmentation by U-Net with Dilation Convolution'. arXiv, Dec. 26, 2019. Accessed: Aug. 31, 2022. [Online]. Available: http://arxiv.org/abs/1912.11947

[36] K. Wickstrøm, M. Kampffmeyer, and R. Jenssen, 'UNCERTAINTY MODELING AND INTERPRETABILITY IN CONVOLUTIONAL NEURAL NETWORKS FOR POLYP SEGMENTATION', in *2018 IEEE 28th International Workshop on Machine Learning for Signal Processing (MLSP)*, Sep. 2018, pp. 1–6. doi: 10.1109/MLSP.2018.8516998.

[37] D.-P. Fan *et al.*, 'PraNet: Parallel Reverse Attention Network for Polyp Segmentation', in *Medical Image Computing and Computer Assisted Intervention – MICCAI 2020*, Cham, 2020, pp. 263–273. doi: 10.1007/978-3-030-59725-2_26.
VOLUME XX, 2022    10